%% file: paper.tex
\documentclass[runningheads]{llncs}

\usepackage{epsfig}
\usepackage{graphicx}
\usepackage{amsmath}
\usepackage{amssymb}
\usepackage{multirow}
\usepackage{subfigure}
\usepackage{enumitem}
\usepackage{colortbl}
\usepackage[usenames,dvipsnames]{xcolor}
\usepackage{xcolor,colortbl}
\usepackage{xfrac}
\usepackage{dblfloatfix}
\usepackage{flushend}
\usepackage{verbatim}
\usepackage{booktabs}
\usepackage{adjustbox}
\usepackage{wrapfig}
\usepackage{microtype}
\usepackage{makecell}
\usepackage{bbm}

\usepackage{xcolor}
\usepackage{textcomp}
\usepackage{color}
\usepackage{colortbl}
\usepackage{url}

\definecolor{yellow}{rgb}{1,1,0.8}
\definecolor{green}{rgb}{0.7,0.95,0.65}
\definecolor{gray}{rgb}{0.9,0.9,0.9}
\definecolor{lightyellow}{rgb}{1,1, 0.8}

\input{macros.tex}

\usepackage[width=122mm,left=22mm,paperwidth=166mm,height=193mm,top=22mm,paperheight=232mm]{geometry}

\begin{document}
\pagestyle{headings}
\mainmatter
\def\ECCVSubNumber{3651}  %

\title{What Matters in Unsupervised Optical Flow}

\author{Rico Jonschkowski$^{1,2}$, Austin Stone$^{1,2}$, Jonathan T. Barron$^{2}$,\\Ariel Gordon$^{1,2}$, Kurt Konolige$^{1,2}$, and Anelia Angelova$^{1,2}$}
\institute{$^1$Robotics at Google and $^2$Google AI\\\email{\{rjon,austinstone,barron,gariel,konolige,anelia\}@google.com}}
\authorrunning{R. Jonschkowski et al.}

\maketitle

\begin{abstract}
We systematically compare and analyze a set of key components in unsupervised optical flow to identify which photometric loss, occlusion handling, and smoothness regularization is most effective. Alongside this investigation we construct a number of novel improvements to unsupervised flow models, such as cost volume normalization, stopping the gradient at the occlusion mask, encouraging smoothness before upsampling the flow field, and continual self-supervision with image resizing. By combining the results of our investigation with our improved model components, we are able to present a new unsupervised flow technique that significantly outperforms the previous unsupervised state-of-the-art and performs on par with supervised FlowNet2 on the KITTI 2015 dataset, while also being significantly simpler than related approaches.
\end{abstract}

\section{Introduction}

Optical flow is a key representation in computer vision that describes the pixel-level correspondence between two images. Since optical flow is useful for estimating motion, disparity, and semantic correspondence, improvements in optical flow directly benefit downstream tasks such as visual odometry, stereo depth estimation, and object tracking.
The performance of optical flow techniques has recently seen dramatic improvements, due to the widespread adoption of deep learning.
Because ground-truth labels for dense optical flow are difficult to obtain for real image pairs, supervised optical flow techniques are primarily trained using synthetic data~\cite{FlowNet}.
Although models trained on synthetic data often generalize well to real images, there is an inherent mismatch between these two data sources that those approaches may struggle to overcome~\cite{Mayer2018,datasetbias}

Though non-synthetic data for training \emph{supervised} optical flow techniques is scarce, the data required to train an \emph{unsupervised} model is abundant: all that training requires is unlabeled video, of which there are countless hours freely available on the internet.
If an unsupervised approach could leverage this abundant and diverse real data, it would produce an optical flow model that does not suffer from any mismatch between its training data and its test data, and could presumably produce higher-quality results.
The core assumption shared by unsupervised optical flow techniques is that an object's appearance does not change as it moves, which allows these models to be trained using unlabeled video as follows: The model is used to estimate a flow field between two images, that flow field is used to warp one image to match the other, and then the model weights are updated so as to minimize the difference between those two images -- and to accommodate some form of regularization. 

Although all unsupervised optical flow methods share this basic idea, their details vary greatly. In this work we systematically compare, improve, and integrate key components to further our understanding and provide a unified framework for unsupervised optical flow. Our contributions are:
\begin{enumerate}[leftmargin=*,itemindent=0pt,parsep=0pt,topsep=0pt]
\item We systematically compare key components of unsupervised optical flow, such as photometric losses, occlusion estimation techniques, self-supervision, and smoothness constraints, and we analyze the effect of other choices, such as pretraining, image resolution, data augmentation, and batch size.
\item We propose four improvements to these key components: cost volume normalization, gradient stopping for occlusion estimation, applying smoothness at the native flow resolution, and image resizing for self-supervision.
\item We integrate the best performing improved components in a unified framework for unsupervised optical flow (UFlow for short) that sets a new state of the art -- even compared to substantially more complex methods that estimate flow from multiple frames or co-train flow with monocular or stereo depth estimation. To facilitate future research, our source code is available at \url{https://github.com/google-research/google-research/tree/master/uflow}.
\end{enumerate}

\section{Related Work}

The motion between an object and a viewer causes apparent movement of brightness patterns in the image~\cite{Gibson1950}. Optical flow techniques attempt to invert this relationship to recover a motion estimate~\cite{lucas1981iterative}. Classical methods infer optical flow for a pair of images by minimizing a loss function that measures photometric consistency and smoothness~\cite{Horn1981,Brox04,Sun2010}. Recent approaches reframe optical flow estimation as a learning problem in which a CNN-based model regresses from a pair of images to a flow field~\cite{FlowNet,Flownet2}. Some models incorporate ideas from earlier methods, such as cost volumes and coarse-to-fine warping~\cite{spynet2017,Sun2018PWCNet,yang2019volumetric}. These supervised approaches require representative training data with accurate optical flow labels. Though such data can be generated for rigid objects with known geometry~\cite{Geiger2012CVPR,KITTI2015}, recovering this ground truth flow for arbitrary scenes is laborious, and requires approaches as unusual as manually painting scenes with textured fluorescent paint and imaging it under ultraviolet light~\cite{Baker2011}. Since such approaches scale poorly, supervised methods have mainly relied on synthetic data for training, and often for evaluation \cite{Barron94performanceof,ButlerECCV2012,FlowNet}. Synthesizing ``good'' training data (such that learned models generalize to real images) is itself a hard research problem, requiring careful consideration of scene content, camera motion, lens distortion, and sensor degradation~\cite{Mayer2018}.

Unsupervised approaches circumvent the need for labels by optimizing photometric consistency with some regularization~\cite{ren2017unsupervised,jjyu2016unsupflow}, similar to the classical optimization-based methods mentioned above. Where traditional methods solve an optimization problem for each image pair, unsupervised learning jointly optimizes an objective across all pairs in a dataset and learns a function that regresses a flow field from images. This approach has two advantages: 1) inference is fast because optimization is only performed during training, and 2) by jointly optimizing across the whole train set, information is shared across image pairs which can potentially improve performance. This unsupervised approach was extended to use edge-aware smoothness~\cite{wang2018occlusion}, a bi-directional Census loss~\cite{meister2018unflow}, different forms of occlusion estimation~\cite{Janai2018ECCV,meister2018unflow,wang2018occlusion}, self-supervision~\cite{DDFlow,SelFlow}, and estimation from multiple frames~\cite{Janai2018ECCV,SelFlow}. Other extensions introduced geometric reasoning through epipolar constraints~\cite{Zhong2019UnsupervisedDE} or by co-training optical flow with depth and ego-motion models from monocular~\cite{ranjan2019cvpr,yin2018geonet,zou2018dfnet} or stereo input~\cite{wang2018unos}.

These works have pushed the state of the art and generated a range of ideas for unsupervised optical flow. But since each of them evaluates a different \emph{combination} of ideas, it is unclear how \emph{individual} ideas compare to each other and which ideas combine well together. For example, the methods OAFlow~\cite{wang2018occlusion} and DDFlow~\cite{DDFlow} use different photometric losses and different ways to mask occlusions, and OAFlow uses an edge-aware smoothness loss while DDFlow regularizes learning through self-supervision. DDFlow performs better than OAFlow, but does this mean that every component of DDFlow is better than every component of OAFlow?
The ablation studies often presented in these papers show that each novel contribution of each work does indeed improve the performance of each individual model, but they do not provide a guarantee that each such contribution will always improve performance when added to any \emph{other} model.
Our work addresses this problem by systematically comparing and combining photometric losses (L1, Charbonnier~\cite{Sun2010}, Census~\cite{DDFlow,meister2018unflow,Zhong2019UnsupervisedDE,zou2018dfnet}, and structural similarity~\cite{ranjan2019cvpr,wang2018unos,yin2018geonet}), different methods for occlusion estimation~\cite{Brox04,wang2018occlusion}, first order and second order edge-aware smoothness~\cite{tomasi1998bilateral}, and self-supervision~\cite{DDFlow}. Our work also improves cost volume computation, occlusion estimation, smoothness, and self-supervision and integrates all components into an state of the art framework for unsupervised optical, while being simpler than many proposed methods to form a solid base for future work.

\section{Preliminaries on Unsupervised Optical Flow}

The task of estimating optical flow can be defined as follows: Given two color images $I^{(1)}, I^{(2)} \in \mathbb{R}^{H \times W \times 3}$, we want to estimate the flow field $V^{(1)}\in \mathbb{R}^{H \times W \times 2}$, which for each pixel in $I^{(1)}$ denotes the relative position of its corresponding pixel in $I^{(2)}$. Note that optical flow is an asymmetric representation of pixel motion: $V^{(1)}$ provides a flow vector for each pixel in $I^{(1)}$, but to find a mapping from image 2 back to image 1, one would need to estimate $V^{(2)}$.

In the context of unsupervised learning, we want to find a function $V^{(1)}=f_\theta(I^{(1)}, I^{(2)})$ with parameters $\theta$ learned from a set of image sequences $D=\{ (I^{(1)}, I^{(2)}, \dots, I^{(N)})\}$. Because we lack ground truth flow, we must define a proxy objective $\lossfun{}(D, \theta)$, such as photometric consistency between $I^{(1)}$ and $I^{(2)}$ after it has been warped according to some estimated $V^{(1)}$. To enforce photometric consistency only for pixels that can be reconstructed from the other image, we must also estimate an occlusion mask $O^{(1)} \in \mathbb{R}^{H \times W}$, for example based on the estimated forward and backward flow fields $O^{(1)}=g(V^{(1)}, V^{(2)})$. $\lossfun{}(\cdot)$ might also include other terms for, e.g. for smoothness or self-supervision. If $\lossfun{}(\cdot)$ is differentiable with respect to $\theta$, the parameters that minimize this loss $\theta^*=\arg\min(\lossfun{}(D, \theta))$ can be recovered using gradient-based optimization.

\section{Key Components of Unsupervised Optical Flow}

This section compares and improves key components of unsupervised optical flow. We will first discuss a model $f_\theta(\cdot)$, which we base on PWC-Net~\cite{Sun2018PWCNet}, and improve through \emph{cost-volume normalization}. Then we go through different components of the objective function $\lossfun{}(\cdot)$: occlusion-aware photometric consistency, smoothness, and self-supervision. Here, we propose improvements to each component: \emph{stopping the gradient at the occlusion mask}, \emph{computing smoothness at the native flow resolution}, and \emph{image resizing for self-supervision}. We end this section by discussing data augmentation and optimization.

\begin{figure}[t]
    \centering
    \includegraphics[width=0.59\columnwidth]{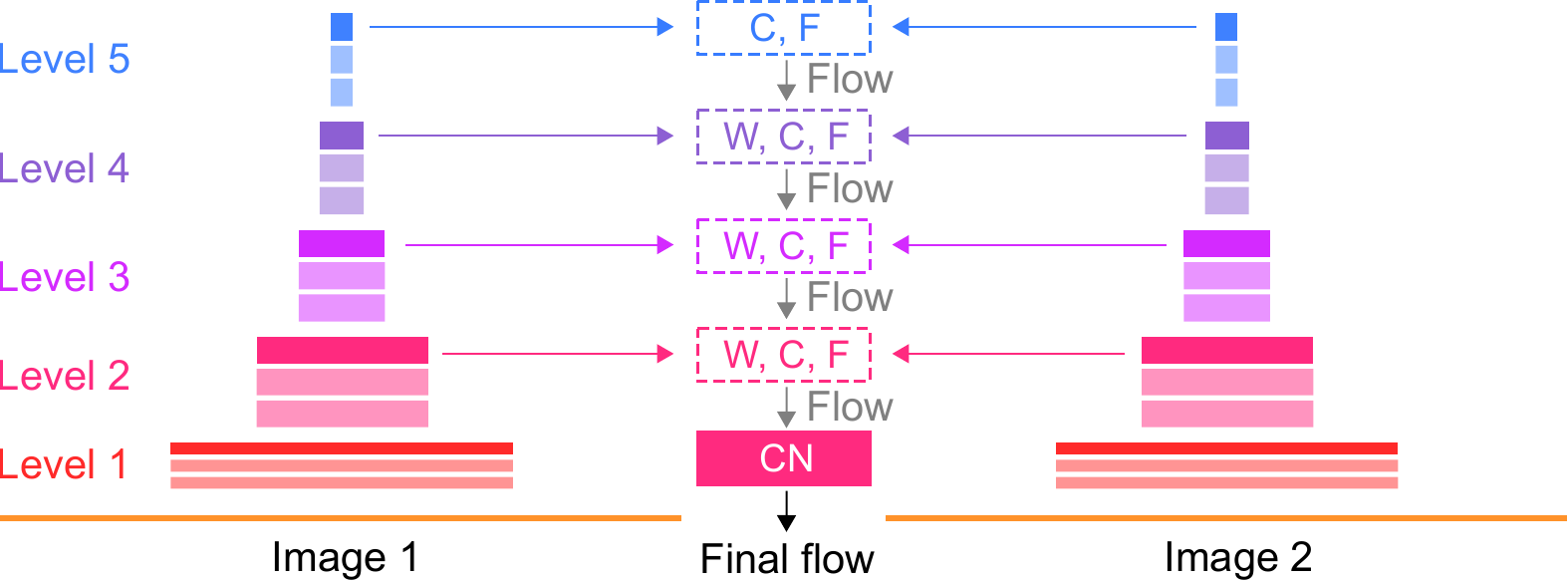}%
    \hspace{0.1cm}
    \includegraphics[width=0.36\columnwidth]{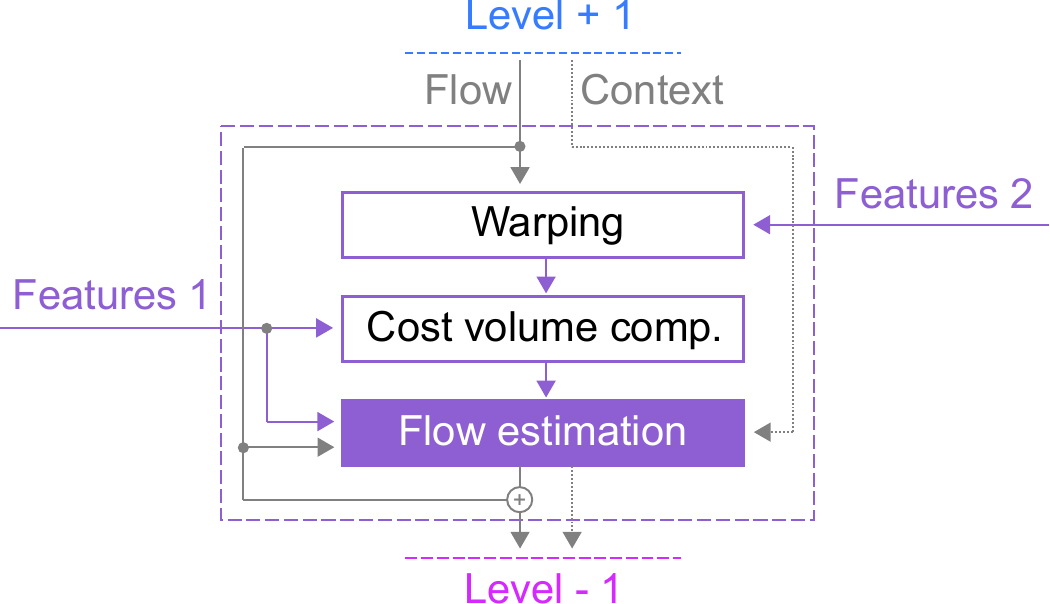}%
    \caption{Model overview. \emph{Left:} Feature pyramids feed into a top-down flow estimation. \emph{Right}: A zoomed in view on a ``W, C, F'' (warping, cost volume, flow estimation) block}
    \label{fig:model}
\end{figure}

As shown in Fig.~\ref{fig:model}, our model feeds images $I^{(1)}$ and $I^{(2)}$ into a shared CNN that generates a feature pyramid, where features are used as input for warping (W), cost volume computation (C), and flow estimation (F). At each level $\level$, the estimated flow $V^{(1, \level+1)}$ from the level above is upscaled, passed down to the lower level as $\hat{V}^{(1, \level+1)}$, and then used to warp $F^{(2, \level)}$, the features of image 2. The warped features $w(F^{(2, \level)}, \hat{V}^{(1, \level)})$ together with $F^{(1, \level)}$ are used to compute a cost volume.
The cost volume considers feature correlations for all pixels and all 81 combinations of shifting $w(F^{(2, \level)}, \hat{V}^{(1, \level)})$ up to 4 pixels up/down and left/right.
This results in a cost volume $C^{\level}\in\mathbb{R}^{\frac{W}{2^\level} \times \frac{H}{2^\level} \times 81}$ that describes how closely each pixel in $F^{(1, \level)}$ resembles the 81 pixels around its location in $F^{(2, \level)}$.
The cost volume, the features from image 1, the higher level flow and the \emph{context} -- the output of the second to last layer of the flow estimation network -- are fed into a CNN that estimates a flow $V^{(1, \level)}$.
After a number of flow estimation levels, there is a final stage of flow refinement at level two in which the flow and context are fed into a \emph{context network} (CN), which is a stack of dilated convolutions.

\paragraph{Model Shrinking, Level Dropout and Cost Volume Normalization:} PWC-Net was designed for supervised learning of optical flow~\cite{Sun2018PWCNet}. To deal with increased memory requirements for unsupervised learning due to bi-directional losses, occlusion estimation, and self-supervision, we remove level six, use 32 channels in all levels, and add residual connections to all flow estimation modules (the ``+'' in the bottom right of Fig.~\ref{fig:model}). Additionally, we dropout residual flow estimation at all levels to further regularize learning, i.e. we randomly pass the resized and rescaled flow estimate from the level above directly to the level below.

Another difference when using this model for unsupervised rather than supervised learning is that unsupervised losses are typically only imposed on the final output (presumably because photometric consistency and other objectives work better at higher resolutions). But without supervised losses on intermediate flow predictions, the model has difficulty learning flow estimation at higher levels. We found that this is caused by very low values in the estimated cost volumes as a result of vanishing feature activations at higher levels.

We address this problem by \emph{cost volume normalization}. Let us denote features for image $i$ at level $\level$ as $F^{(i,\level)} \in \mathbb{R}^{\frac{H}{2^\level}\times \frac{W}{2^\level}\times d}$. The cost volume between images $1$ and $2$ for all image locations $(x, y)$ and all considered image shifts $(u, v)$ is the inner product of the normalized features of the two images:
\begin{equation}
C^{(\level)}_{x, y, u,v} = \sum_d \left(\frac{ F^{(1, \level)}_{x, y, d} - \mu^{(1, \level)}}{\sigma^{(1, \level)}} \right) \left( \frac{F^{(2,\level)}_{x+u, y+v, d}  - \mu^{(2, \level)}}{\sigma^{(2, \level)}} \right) \, .
\end{equation}
Where $\mu^{(i, \level)}$ and $\sigma^{(i, \level)}$ are the sample mean and standard deviation of $F^{(i,\level)}$ over its spatial and feature dimensions. We found that cost volume normalization improves convergence and final performance in unsupervised optical flow. These findings are consistent with prior work that used a similar form of normalization to improve geometric matching~\cite{rocco2017convolutional}.

\subsubsection{Unsupervised Learning Objectives:} Defining a learning objective $\lossfun{}(\cdot)$ that specifies the task of learning optical flow without having access to labels is the core problem of unsupervised optical flow. Similar to related work~\cite{DDFlow,SelFlow,meister2018unflow,wang2018occlusion}, we train our model by estimating optical flow and applying the respective losses in both directions. In this work we consider a learning objective that consists of three terms: occlusion-aware photometric consistency, edge-aware smoothness, and self-supervision, which we will now discuss in detail.  

\noindent{\emph{Photometric Consistency}:} The photometric consistency term encourages the estimated flow to align image patches with a similar appearance by penalizing photometric dissimilarity. The metric for measuring appearance similarity is critical for any unsupervised optical flow technique. Related approaches use three different objectives here (sometimes in combination), (i) the generalized Charbonnier loss~\cite{Janai2018ECCV,meister2018unflow,ren2017unsupervised,wang2018occlusion,jjyu2016unsupflow,Zhong2019UnsupervisedDE}, (ii) the structural similarity index (SSIM) loss~\cite{yin2018geonet,ranjan2019cvpr,wang2018unos}, and (iii) the Census loss~\cite{meister2018unflow,Zhong2019UnsupervisedDE,zou2018dfnet}. We compare all three losses in this paper.
The generalized Charbonnier loss~\cite{Sun2010} is $\lossfun{C}=\frac{1}{n}\sum\left((I^{(1)} - w(I^{(2)})^2 + \epsilon^2 \right)^\alpha$. Our experiments use $\epsilon=0.001$ and $\alpha=0.5$ and also compare to using a modified L1 loss $\lossfun{L1}=\sum|I^{(1)} - w(I^{(2)}) + \epsilon'|$ with $\epsilon'=10^{-6}$.
For the SSIM~\cite{Wang04imagequality} loss, we use an occlusion-aware implementation from recent work~\cite{Gordon2019}. For the Census loss, we use a soft Hamming distance on Census-transformed image patches~\cite{meister2018unflow}. Based on the empirical results discussed below, we use the Census loss unless otherwise stated. All photometric losses are computed using an occlusion-masked average over all pixels~\cite{wang2018occlusion}. 

\noindent{\emph{Occlusion Estimation}:} By definition, occluded regions do not have a correspondence in the other image, so they should be discounted when computing the photometric loss.
Related approaches estimate occlusions by (i) checking for consistent forward and backward flow~\cite{wang2018occlusion}, (ii) using the range map of the backward flow~\cite{Brox04}, and (iii) learning a model for occlusion estimation~\cite{Janai2018ECCV}. We are considering and comparing the first two variants here and improve the second variant through gradient stopping. In addition to taking into account occlusions, we also mask ``invalid'' pixels whose flow vectors point outside of the frame of the image~\cite{wang2018occlusion}.
The forward-backward consistency check defines occlusions a pixels for which the flow and the back-projected backward flow disagree by more than a threshold, such that the occlusion mask is defined as $O^{(1)} = \mathbbm{1}_{|V^{(1)} - w(V^{(2)})|^2 < \alpha_1(|V^{(1)}|^2 - |w(V^{(2)})|^2) + \alpha_2}$, where $\alpha_1=0.01$ and $\alpha_2=0.5$~\cite{sundaram2010dense}. An alternative approach computes a ``range map'' $R^{(i)} \in \mathbb{R}^{H \times W}$ -- a soft histogram of how many pixels in the other image map onto a given pixel, which is constructed by having each flow vector distribute a total weight of 1 to the four pixels around its end point according to a bilinear kernel~\cite{wang2018occlusion}. Pixels that none of the reverse flow vectors point to are assumed to have no correspondence in the other image, and are therefore occluded. As proposed by Wang~\etal~\cite{wang2018occlusion}, we compute an occlusion mask $O^{(i)}\in\mathbb{R}^{W \times H}$ by thresholding the range map at 1. Based on the empirical results below, we use range-map based occlusion estimation by default, but use the forward-backward consistency check on KITTI, where it significantly improves performance.

\noindent{\emph{Gradient Stopping at Occlusion Masks}:} Although prior work does not mention this issue~\cite{wang2018occlusion}, we found that propagating the gradient of the photometric loss into the occlusion estimation consistently degraded performance or caused divergence when the occlusion estimation was differentiable, as is the case for range-map based occlusion. This behavior is to be expected because when computing the occlusion-weighted average over photometric dissimilarity, there should be a gradient towards masking pixels with high photometric error. We address this problem by stopping the gradient at the occlusion mask, which eliminates divergence and improves performance.

\noindent{\emph{Smoothness}:} Different forms of smoothness are commonly used to regularize optical flow in traditional methods~\cite{Brox04,Horn1981,Sun2010} as well as most recent unsupervised approaches~\cite{jjyu2016unsupflow,ren2017unsupervised,wang2018occlusion,meister2018unflow,yin2018geonet,zou2018dfnet,ranjan2019cvpr,Janai2018ECCV,wang2018unos,Zhong2019UnsupervisedDE}. In this work, we consider edge-aware first and second order smoothness~\cite{tomasi1998bilateral}, where flows are encouraged to align their boundaries with visual edges in the image $I^{(1)}$. Formally, we define $k$th order smoothness as:
\begin{equation}
\resizebox{0.93\linewidth}{!}{$
\lossfun{smooth(k)} =
\mean{ \exp {\left(- \frac{\lambda}{3} \sum_c \abs{\frac{\partial I_c^{(1, \level)}}{\partial x}}\right)} \abs{ \frac{\partial^k V^{(1, \level)}}{\partial x^k} } + \exp {\left(-\frac{\lambda}{3}\sum_c\abs{\frac{\partial I_c^{(1, \level)}}{\partial y}}\right)} \abs{\frac{\partial^k V^{(1, \level)}}{\partial y^k}}}\,.
$}
\label{eq:smooth}
\end{equation}
Where $\lambda$ modulates edge weighting based on $I^{(1, \level)}_c$ for color channel $c\in[0,2]$. By default, we use first order smoothness on Flying Chairs and Sintel and second order smoothness on KITTI, which we ablate in different experiments.

\noindent{\emph{Smoothness at Flow Resolution}:} A question that we have not seen addressed is at which level $\level$, smoothness should be applied. Since we follow the commonly used method of estimating optical flow at $\level=2$, i.e. at a quarter of the input resolution, followed by upsampling through bilinear interpolation, our model produces piece-wise linear flow fields. As a result, only every fourth pixel can possibly have a non-zero second order derivative, which might not be aligned with the corresponding image edge and thereby reduce the effectiveness of edge-aware smoothness. To address this, we apply smoothness at level $\level=2$ where flow is generated and downsample the image instead of upsampling the flow. This of course does not affect evaluation, which is done at the original image resolution.

\noindent{\emph{Self-supervision}:} The idea of self-supervision in unsupervised optical flow is to generate optical flow labels by applying the learned model on a pair of images, then modify the images to make flow estimation more difficult and train the model to recover the originally estimated flow~\cite{DDFlow,SelFlow}. Since we see the main utility of this technique in learning flow estimation for pixels that go out of the image boundary -- where cost-volume computation is not informative and photometric losses do not apply -- we build on and improve ideas about self-supervised image crops~\cite{DDFlow}. 
For our self-supervised objective, we apply our model on the full images, crop the images by removing 64 pixels from each edge, apply the model again, and use the cropped estimated flow from the full images as supervision for flow estimation from the cropped images. We define the self-supervision objectives as an occlusion-weighted Charbonnier loss, that takes into account only pixels that have low forward-backward consistency in the ``student'' flow from cropped image and high forward-backward consistency in the ``teacher'' flow from the original images, similar to DDFlow~\cite{DDFlow}.

\noindent{\emph{Continual Self-supervision and Image Resizing}:} Unlike related work, we do not first train and then freeze a teacher model to supervise a separate student model but rather have a single model that supervises itself, which simplifies the approach, reduces the required memory, and allows the self-supervision signal to improve continually. To stabilize learning, we stop gradients of the self-supervision objectives to be propagated into the ``teacher'' flow. Additionally, we resize the image crops to match the original resolution before feeding them into the model (and we rescale the self-generated flow labels accordingly) to make the self-supervision examples more representative of the problem of extrapolating flow beyond the image boundary in the original size.

\subsubsection{Optimization:}

To train our model $f_\theta(\cdot)$ we minimize a weighted sum of losses:
\begin{equation}
\lossfun{}(D, \theta) = w_{\mathit{photo}} \cdot \lossfun{photo} + w_{\mathit{smooth}} \cdot \lossfun{smooth} + w_{\mathit{self}} \cdot \lossfun{self},
\end{equation}
where $\lossfun{photo}$ is our photometric loss, $\lossfun{smooth}$ is smoothness regularization, and $\lossfun{self}$ is the self-supervision Charbonnier loss. 
We set $w_{\mathit{photo}}$ to $1$ for experiments using the Census loss and to $2$ when we compare to the SSIM, Charbonnier, or L1 losses. We set $w_{\mathit{self}}$ to $2$ when using first order, and to $4$ for second order smoothness and use an edge-weight of $\lambda=150$. We use $w_{\mathit{self}}=0$ during the first half of training, linearly increase it $0.3$ during the next $10\%$ of gradient steps and keep it constant afterwards.

RGB image values are scaled to $[-1, 1]$, and augmentated by randomly swapping the color channels and randomly shifting the hue. Sintel images are additionally randomly flipped up/down and left/right. All models are trained using with Adam~\cite{KingmaAdam} ($\beta_1=0.9$, $\beta_2=0.999$, $\epsilon=10^{-8}$) with a learning rate of $10^{-4}$ for $m$ steps, followed by another $\frac{1}{5}m$ steps during which the learning rate is exponentially decayed to $10^{-8}$. All ablations use $m=\text{50K}$ with batch size 32, but the final model was trained using $m=\text{1M}$ with batch size 1, which produced slightly better performance as described below. Either way, the training takes about three days. Experiments on Sintel and KITTI start from a model that was first trained on Flying Chairs.

\section{Experiments}
\label{sec:experiments}

We evaluate our model on the standard optical flow benchmark datasets: Flying Chairs \cite{FlowNet}, Sintel \cite{ButlerECCV2012}, and KITTI 2012/2015 \cite{Geiger2012CVPR,KITTI2015}. We divide Flying Chairs and Sintel according to its standard train/test split. For KITTI, we train on the multi-view extension on the KITTI 2015 dataset, and we do not train on any data from KITTI 2012 because it does not have moving objects.

Related work is inconsistent in their use of train/test splits. For Sintel, it is common to train on the training set, report the benchmark performance on the test set, and evaluate ablations on the training set only (because test set labels are not public), which does not test generalization very well.
Others ``download the Sintel movie and extract $\sim\!$10,000 images''~\cite{SelFlow} including the test set images, which is intended to demonstrate the ability of unsupervised methods to train on raw video data, but unfortunately also includes the benchmark test images in the training set.
For KITTI, other works train on the raw KITTI dataset with and without excluding the evaluation set, or most commonly train on frames 1--8 and 13--20 of the multi-view extension of KITTI 2012/2015 datasets and evaluate on frames 10/11.
But this split can mask overfitting to the trained sequences -- either in the ablation results or also in the benchmark results, when the multiview-extensions of both the train and the test set are used. We therefore adopt the training regimen of Zhong \etal~\cite{Zhong2019UnsupervisedDE} and train two models for each dataset, one on the training set and one on test set (or for KITTI on their multiview extension) and evaluate these models appropriately.

Following the conventions of the KITTI benchmark, we report endpoint error (``EPE'') and error rates (``ER''), where a prediction is considered erroneous if its EPE is $>3$ pixels and if the distance between the predicted point and the true end point is $>5\%$ of the length of the true flow vector. We compute these metrics for all pixels (``occ'' in the KITTI benchmark, which we call ``all'' in this paper).
We use the common practice of pretraining on the train split of the Flying Chairs dataset before training on Sintel / KITTI. We evaluate on all images in the native resolution, but have the model perform inference on a resolution that is divisible by 32, output at a four times smaller resolution, and then resize the output to the original resolution for evaluation. On KITTI, we observe that performance improves when using a square input resolution instead of a resolution in the original aspect ratio -- perhaps because KITTI is dominated by horizontal motion. Accordingly, we use the following resolutions in our experiments: Flying Chairs: 384$\times$512, Sintel: 448$\times$1024, KITTI: 640$\times$640.

\section{Results}

We evaluate our model in an extensive comparison and ablation study, from which we identify the best combination of components, tested in the ``full'' setting, which is often different from the components that work best individually in our ``minimal'' setting (more details below). We then compare our resulting model to the best published methods on unsupervised optical flow, and show that it outperforms all methods on all benchmarks.

\subsubsection{Ablations and Comparisons of Key Components}
To determine which aspects of unsupervised optical flow are most important, we perform an extensive series of ablation studies.
We find that a) occlusion-masking, self-supervision, and smoothness are all important, b) level dropout and cost volume normalization improve performance, c) the Census loss outperforms other photometric losses, d) range-map based occlusion estimation requires gradient stopping to work, c) edge-aware smoothness and smoothness level matters significantly, d) self supervision helps especially for KITTI, and is improved by our changes, e) losses might be the current performance bottleneck, f) changing the resolution can substantially improve results, g) data augmentation and pretraining are helpful.

In each ablation study we train one model per domain (on Flying Chairs, KITTI-test, and Sintel-test), and evaluate those on the corresponding validation split from the same domain, taking into account occluded and non-occluded pixels ``(all)''. To estimate the noise in our results, we trained models with six different random seeds for each domain and computed their standard deviations per metric: Flying Chairs: 0.0162, Sintel Clean: 0.0248, Sintel Final: 0.0131, KITTI-2015: 0.0704, 0.0718\%. We now describe the findings of each study.

\begin{wraptable}{r}{0.45\columnwidth}
     \vspace{-10pt}
    \caption{Core components: OM: occlusion masking, SM: smoothness, SS: self-supervision; ``div.'': divergence}
    \label{table:core_components}
    \adjustbox{width=\linewidth}{
    \begin{tabular}{ccc ccccccccc}
    \toprule
    &&&\phantom{a}& Chairs & \phantom{a} & \multicolumn{2}{c}{Sintel \textit{train}} & \phantom{a} & \multicolumn{3}{c}{KITTI-15 \textit{train}} \\
    \cmidrule{5-5} \cmidrule{7-8} \cmidrule{10-12} 
    OM & SM & SS && \textit{test} && Clean & Final && all & noc & ER\%\\
    \midrule
    -- & -- & -- && 3.58 && 4.20 & 6.80 && 13.07 & 2.47 & 21.21 \\
    -- & -- & \checkmark && 2.99 && 3.34 & 5.18 && 11.36 & 2.30 & 18.61 \\
    -- & \checkmark & -- && 2.84 && 3.37 & 5.19 && 11.37 & 2.17 & 19.31 \\
    -- & \checkmark & \checkmark && 2.74 && 3.12 & 4.56 && 3.28 & 2.08 & 9.97 \\
    \checkmark & -- & -- && 3.28 && 3.78 & 5.85 && div. & div. & div. \\
    \checkmark & -- & \checkmark && 2.91 && 3.26 & 4.72 && 3.02 & 2.11 & 9.89 \\
    \checkmark & \checkmark & -- && 2.63 && 3.20 & 4.63 && 4.15 & 2.05 & 13.15 \\
    \checkmark & \checkmark & \checkmark && {\bf 2.55} && {\bf 3.00} & {\bf 4.18} && {\bf 2.94} & {\bf 1.98} & {\bf 9.65} \\
    \bottomrule
    \end{tabular}}
\end{wraptable}

\noindent{\emph{Core Components}:} Table~\ref{table:core_components} shows how performance varies as each core component of our model (occlusion masking, smoothness, and self-supervision) is removed. We see that every component contributes to the overall performance. Since the utility of different components depends on what other components are used, all following experiments compare to the ``minimal'' (first row) and ``full'' (last row) versions of our method. Qualitative results for rows 9, 4, 3, and 1 are shown in Figure~\ref{fig:ablation} (from left to right). Note how the flow error $\Delta V$ increases with each removal of a core component.

\pagebreak

\begin{wraptable}{r}{0.45\columnwidth}
     \vspace{-10.5pt}
    \caption{Model improvements. CVN: cost volume normalization, LD: level dropout}
    \label{table:model}
    \adjustbox{width=\linewidth}{
    \begin{tabular}{lcccccccccccc}
    \toprule
    &\phantom{a}&&&\phantom{a}& Chairs & \phantom{a} & \multicolumn{2}{c}{Sintel \textit{train}} & \phantom{a} & \multicolumn{3}{c}{KITTI-15 \textit{train}} \\
    \cmidrule{6-6} \cmidrule{8-9} \cmidrule{11-13} 
    && \phantom{a}CVN\phantom{a} & \phantom{a}LD\phantom{a} && \textit{test} && Clean & Final && all & noc & ER\%\\
    \midrule
    \parbox[t]{2mm}{\multirow{4}{*}{\rotatebox[origin=c]{90}{Minimal}}} 
    && -- & -- && 5.01 && 4.52 & 6.67 && 13.30 & 2.72 & 21.69 \\
    && -- & \checkmark && 5.29 && 4.40 & {\bf6.59} && {\bf12.75} & 2.49 & 21.30 \\
    && \checkmark & -- && 4.86 && {\bf4.19} & 6.69 && 13.294 & 2.59 & 21.54 \\
    && \checkmark & \checkmark && {\bf3.58} && 4.20 & 6.80 && 13.07 & {\bf2.47} & {\bf21.21} \\
    \hline
    \parbox[t]{2mm}{\multirow{4}{*}{\rotatebox[origin=c]{90}{Full}}} 
    && -- & -- && 3.78 && 3.41 & 4.70 && 39.09 & 30.19 & 98.77 \\
    && -- & \checkmark && 3.21 && 3.45 & 4.61 && 2.96 & {\bf 1.96} & 9.77 \\
    && \checkmark & -- && {\bf2.54} && 3.07 & 4.31 && 3.16 & 2.04 & 10.35 \\
    && \checkmark & \checkmark && 2.55 && {\bf 3.00} & {\bf 4.18} && {\bf 2.94} & 1.98 & {\bf 9.65} \\
    \bottomrule
    \end{tabular}}
    \caption{Photometric losses. Best results of L1 and Charbonnier underlined}
    \label{table:photometric_losses}
    \adjustbox{width=\linewidth}{
    \begin{tabular}{llccccccccc}
    \toprule
    &&\phantom{a}& Chairs & \phantom{a} & \multicolumn{2}{c}{Sintel \textit{train}} & \phantom{a} & \multicolumn{3}{c}{KITTI-15 \textit{train}} \\
    \cmidrule{4-4} \cmidrule{6-7} \cmidrule{9-11} 
    &Method && \textit{test} && Clean & Final && all & noc & ER\%\\
    \midrule
    \parbox[t]{2mm}{\multirow{4}{*}{\rotatebox[origin=c]{90}{Minimal}}} 
    & L1 && \underline{4.27} && 5.51 & 7.74 && 17.02 & 6.11 & 32.96 \\
    & Charbonnier && 4.31 && \underline{5.50} & \underline{7.64} && \underline{16.94} & \underline{6.09} & \underline{32.84} \\
    & SSIM && {\bf 3.51} && {\bf 4.01} & {\bf 5.41} && 11.99 & 2.46 & 21.72 \\
    & Census && 3.54 && 4.23 & 6.98 && {\bf 11.66} & {\bf 2.37} & {\bf 21.15} \\
    \hline
    \parbox[t]{2mm}{\multirow{4}{*}{\rotatebox[origin=c]{90}{Full}}} 
    & L1 && \underline{2.83} && \underline{4.23} & \underline{5.75} && \underline{5.53} & \underline{3.17} & \underline{18.65} \\
    & Charbonnier && 2.86 && 4.24 & 5.81 && 5.56 & 3.21 & 18.82 \\
    & SSIM && {\bf 2.54} && 3.08 & 4.52 && 3.29 & 2.04 & 10.41 \\
    & Census && 2.61 && \textbf{3.00} & \textbf{4.20} && {\bf 3.08} & {\bf 2.01} & {\bf 10.01} \\
    \bottomrule
    \end{tabular}}
    \caption{Occlusion estimation. RM: range-map based occllusion, FB: forward-backward consistency check}
    \label{table:occlusion_estimation}
    \adjustbox{width=\linewidth}{
    \begin{tabular}{llccccccccc}
    \toprule
    &&\phantom{a}& Chairs & \phantom{a} & \multicolumn{2}{c}{Sintel \textit{train}} & \phantom{a} & \multicolumn{3}{c}{KITTI-15 \textit{train}} \\
    \cmidrule{4-4} \cmidrule{6-7} \cmidrule{9-11} 
    &Method && \textit{test} && Clean & Final && all & noc & ER\%\\
    \midrule
    \parbox[t]{2mm}{\multirow{5}{*}{\rotatebox[origin=c]{90}{Minimal}}} 
    & None && 3.51 && 4.15 & 6.69 && 12.89 & 2.41 & 21.17 \\
    & RM (w/o grad stop) && div. && div. & div. && div. & div. & div. \\
    & RM (w/ grad stop) && {\bf 3.27} && 3.78 & 5.86 && 10.65 & 2.29 & 18.76 \\
    & FB (from step 1) && 3.57 && {\bf 3.71} & {\bf 4.83} && {\bf 8.99} & 2.16 & {\bf 17.71} \\
    & FB (after 20\% steps) && 3.49 && 3.76 & 4.92 && 9.75 & {\bf 2.13} & 18.38 \\
    \hline
    \parbox[t]{2mm}{\multirow{5}{*}{\rotatebox[origin=c]{90}{Full}}} 
    & None && 2.73 && 3.84 & 5.13 && 3.28 & 2.10 & 10.07 \\
    & RM (w/o grad stop) && div. && div. & div. && div. & div. & div. \\
    & RM (w/ grad stop) && {\bf 2.58} && {\bf 3.01} & 4.25 && 3.10 & 2.04 & 9.86 \\
    & FB (from step 1) && 3.28 && 3.49 & 4.45 && 2.96 & 1.99 & 9.65 \\
    & FB (after 20\% steps)&& 3.14 && 3.12 & {\bf 4.13} && {\bf 2.88} & {\bf 1.95} & {\bf 9.54} \\
    \bottomrule
    \end{tabular}}
\end{wraptable}

\noindent{\emph{Model Improvements}:} Table~\ref{table:model} shows that level dropout (LD) and cost volume normalization (CVN) improve performance in the full setting (but not generally in the minimal setting). CVN appears to be more important for Chairs and Sintel while LD helps most for KITTI.

\noindent{\emph{Photometric Losses}:} Table~\ref{table:photometric_losses} compares commonly used photometric losses and shows that it is important to test every component with the full method, rather than looking at isolated performance. By itself, the commonly-used Charbonnier loss works better, but in the full setting, it underperforms the simpler L1 loss. For KITTI, Census works best in both settings. But for Sintel (in particular Sintel Final), the SSIM loss significantly outperforms Census in the minimal setting (5.41 vs. 6.98) but does not perform as well when used with all components in the full setting.

\newcommand{\resultswidth}{1.06in}
\newcommand{\resultsspace}{\,}
\newcommand{\sidelabel}[1]{\rot{\footnotesize #1}}

\newcommand{\ablationimage}[1]{
\includegraphics[width = \resultswidth]{images/core_components/#1}
}

\begin{figure}[t]
\centering
\resizebox{\linewidth}{!}{
\begin{tabular}{@{}c@{\,\,}c @{}c@{}c@{}c@{\resultsspace}c@{\resultsspace}c@{\resultsspace}c@{}}
\rotatebox{90}{\footnotesize 2015}
&
\raisebox{0.13in}{$I$}
&
\ablationimage{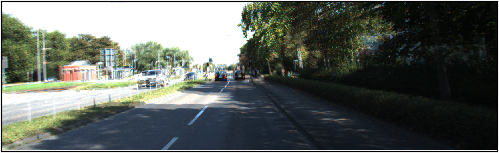}
&
\raisebox{0.13in}{$\Delta V$} & 
\ablationimage{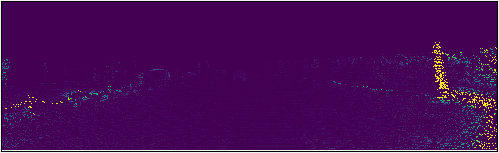} &
\ablationimage{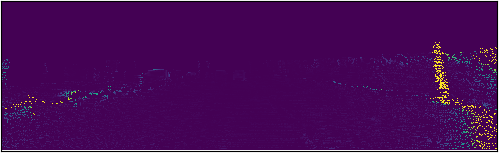} &
\ablationimage{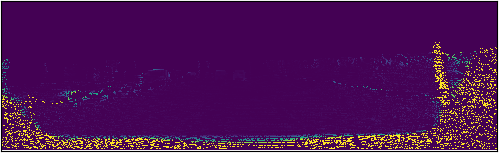} &
\ablationimage{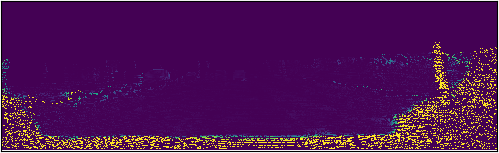} \\
\rotatebox{90}{\footnotesize \!\!KITTI}
&
\raisebox{0.13in}{$V^*$}
&
\ablationimage{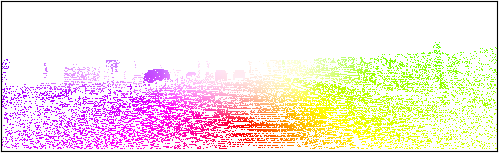} &
\raisebox{0.13in}{$V$}
&
\ablationimage{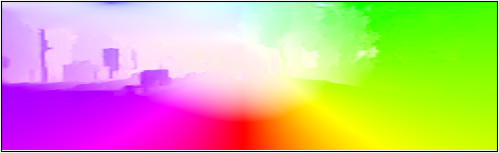} &
\ablationimage{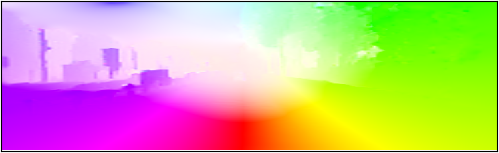} &
\ablationimage{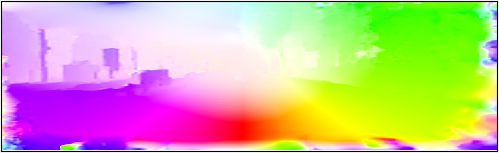} &
\ablationimage{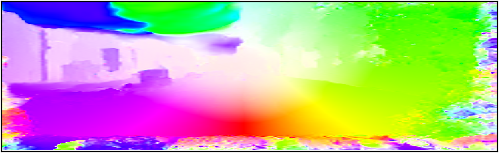} \\
\rotatebox{90}{\footnotesize Final}
&
\raisebox{0.18in}{$I$}
&
\ablationimage{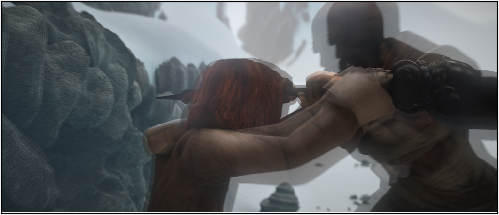}
&
\raisebox{0.18in}{$\Delta V$}
& \ablationimage{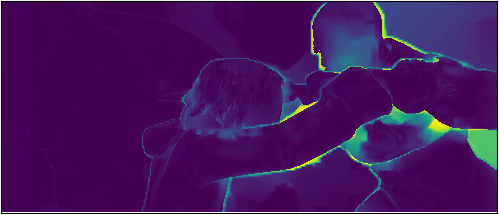} &
\ablationimage{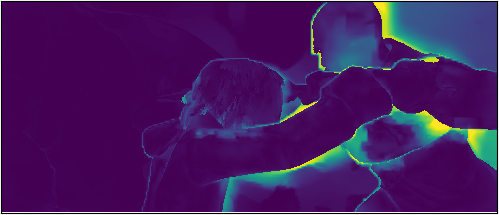} &
\ablationimage{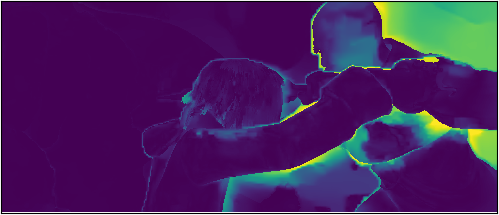} &
\ablationimage{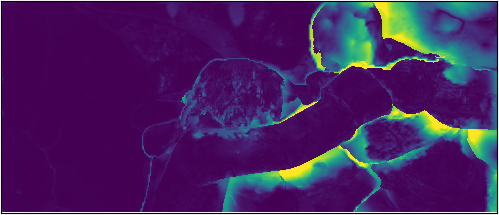} \\
\rotatebox{90}{\footnotesize \,\,\,\,\,\,Sintel}
&
\raisebox{0.18in}{$V^*$}
&
\ablationimage{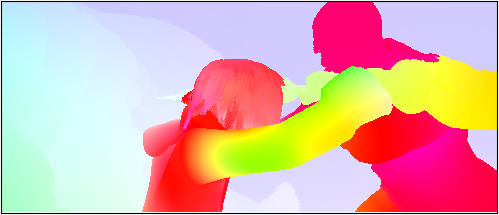}
&
\raisebox{0.18in}{$V$}
&
 \ablationimage{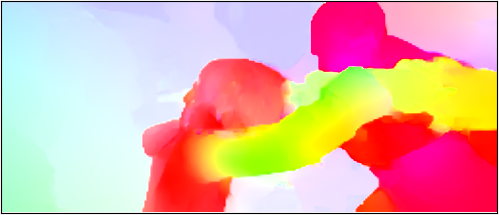} &
\ablationimage{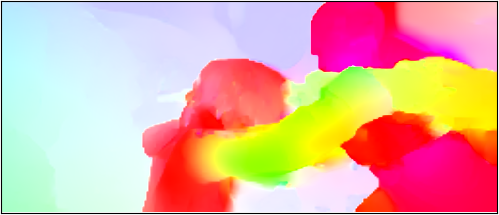} &
\ablationimage{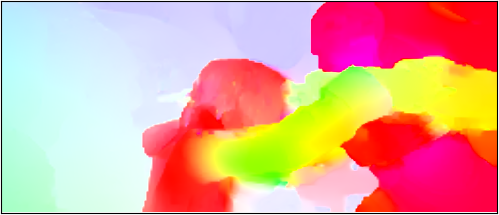} &
\ablationimage{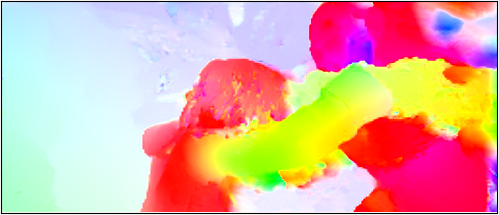} \\
&
&
{\footnotesize Ground truth} && {\footnotesize Full} & {\footnotesize $-$Occlusion} & {\footnotesize $-$Self-supervision} & {\footnotesize $-$Smoothness}
\end{tabular}
}
\caption{Qualitative ablation results of our model on random images not seen during training. Flow quality deteriorates as we progressively ablate core components}
\label{fig:ablation}
\end{figure}

\noindent{\emph{Occlusion Estimation}:} Table~\ref{table:occlusion_estimation} compares different approaches to occlusion estimation (forward-backward consistency and range maps). We see that range-map based occlusion consistently diverges unless we stop the gradient of the photometric loss. But when gradients are stopped, this method works well, especially for Flying Chairs and Sintel Clean. Forward-backward consistency works best for KITTI, especially if not applied from the beginning.

\pagebreak

\begin{wraptable}{r}{0.45\columnwidth}
    \vspace{-10.5pt}
    \caption{Level for smoothness loss}
    \label{table:smoothness_level}
    \adjustbox{width=\linewidth}{
    \begin{tabular}{lcccccccccc}
    \toprule
    &Smoothn.&\phantom{a}& Chairs & \phantom{a} & \multicolumn{2}{c}{Sintel \textit{train}} & \phantom{a} & \multicolumn{3}{c}{KITTI-15 \textit{train}} \\
    \cmidrule{4-4} \cmidrule{6-7} \cmidrule{9-11} 
    &level && \textit{test} && Clean & Final && all & noc & ER\%\\
    \midrule
    \parbox[t]{2mm}{\multirow{3}{*}{\rotatebox[origin=c]{90}{Minimal}}}
    &0 && 3.05 && 4.10 & 5.22 && 12.16 & 2.32 & 20.33 \\
    & 1 && 2.94 && 3.65 & {\bf 5.07} && 11.94 & 2.24 & 19.98 \\
    & 2 && {\bf 2.85} && {\bf 3.33} & 5.21 && {\bf 11.43} & {\bf 2.23} & {\bf 19.38} \\
    \hline
    \parbox[t]{2mm}{\multirow{3}{*}{\rotatebox[origin=c]{90}{Full}}} 
    & 0 && 2.87 && 3.65 & 4.63 && 2.95 & 2.02 & 9.87 \\
    & 1 && 2.74 && 3.13 & 4.29 && 2.96 & {\bf 1.99} & 9.78 \\
    & 2 && {\bf 2.58} && {\bf 3.00} & {\bf 4.24} && {\bf 2.93} & {\bf 1.99} & {\bf 9.63} \\
    \bottomrule
    \end{tabular}}
    \caption{Comparison of weights for first/second order smoothness}
    \label{table:smoothness_weights}
    \adjustbox{width=\linewidth}{
    \begin{tabular}{lccccccccccc}
    \toprule
    &\multicolumn{2}{c}{$w_{\mathit{smooth}}$} &\phantom{a}& Chairs & \phantom{a} & \multicolumn{2}{c}{Sintel \textit{train}} & \phantom{a} & \multicolumn{3}{c}{KITTI-15 \textit{train}} \\
    \cmidrule{5-5} \cmidrule{7-8} \cmidrule{10-12} 
    &1st & 2nd && \textit{test} && Clean & Final && all & occ & ER\%\\
    \midrule
    \parbox[t]{2mm}{\multirow{5}{*}{\rotatebox[origin=c]{90}{Minimal}}}
    & 0 & 0 && 4.55 && 4.16 & 6.84 && div. & div. & div. \\
    & 0 & 2 && 3.13 && 3.77 & 6.32 && 11.37 & 2.17 & 19.33 \\
    & 0 & 8 && 4.02 && 3.50 & 6.08 && {\bf 7.27} & {\bf 2.11} & {\bf 14.70} \\
    & 4 & 0 && {\bf 2.85} && {\bf 3.35} & {\bf 5.05} && 7.23 & 2.30 & 18.58 \\
    & 16 & 0 && 4.37 && 4.78 & 6.03 && 9.58 & 4.09 & 22.82 \\
    \hline
    \parbox[t]{2mm}{\multirow{5}{*}{\rotatebox[origin=c]{90}{Full}}} 
    & 0 & 0 && 2.92 && 3.27 & 4.77 && 2.92 & 2.07 & 9.75 \\
    & 0 & 2 && 2.79 && div. & div. && {\bf 2.93} & 1.98 & {\bf 9.61} \\
    & 0 & 8 && 2.75 && 3.33 & 4.77 && 2.94 & {\bf 1.91} & 9.85 \\
    & 4 & 0 && {\bf 2.60} && {\bf 3.00} & {\bf 4.17} && 5.39 & 2.03 & 16.58 \\
    & 16 & 0 && 3.68 && 4.22 & 5.30 && 8.71 & 4.01 & 21.52 \\
    \bottomrule
    \end{tabular}}
    \caption{Smoothness edge-weights}
    \label{table:smoothness_edge}
    \adjustbox{width=\linewidth}{
    \begin{tabular}{llccccccccc}
    \toprule
    & &\phantom{a}& Chairs & \phantom{a} & \multicolumn{2}{c}{Sintel \textit{train}} & \phantom{a} & \multicolumn{3}{c}{KITTI-15 \textit{train}} \\
    \cmidrule{4-4} \cmidrule{6-7} \cmidrule{9-11} 
    &$\lambda$ && \textit{test} && Clean & Final && all & noc & ER\%\\
    \midrule
    \parbox[t]{2mm}{\multirow{3}{*}{\rotatebox[origin=c]{90}{Minimal}}}
    & 0 && 4.93 && 6.00 & 6.65 && {\bf 4.15} & 2.36 & 12.50 \\
    & 10 && 4.33 && 5.32 & 6.12 && 4.22 & {\bf 2.17} & {\bf 12.28} \\
    & 150 && {\bf 2.83} && {\bf 3.36} & {\bf 5.12} && 11.41 & 2.21 & 19.37 \\
    \hline
    \parbox[t]{2mm}{\multirow{3}{*}{\rotatebox[origin=c]{90}{Full}}} 
    & 0 && 4.87 && 5.78 & 6.40 && 3.86 & 2.84 & 11.81 \\
    & 10 && 3.75 && 4.62 & 5.34 && 3.14 & 2.11 & 10.27 \\
    & 150 && {\bf 2.56} && {\bf 3.02} & {\bf 4.20} && {\bf 2.87} & {\bf 1.95} & {\bf 9.59} \\
    \bottomrule
    \end{tabular}}
    \caption{Self-supervision ablation} 
    \label{table:selfsup}
    \adjustbox{width=\linewidth}{
    \begin{tabular}{llccccccccc}
    \toprule
    &&\phantom{a}& Chairs & \phantom{a} & \multicolumn{2}{c}{Sintel \textit{train}} & \phantom{a} & \multicolumn{3}{c}{KITTI-15 \textit{train}} \\
    \cmidrule{4-4} \cmidrule{6-7} \cmidrule{9-11} 
    &Self-supervision && \textit{test} && Clean & Final && all & noc & ER\%\\
    \midrule
    \parbox[t]{2mm}{\multirow{4}{*}{\rotatebox[origin=c]{90}{Minimal}}}
    & None && 3.48 && 4.10 & 6.62 && 13.05 & 2.48 & 21.23 \\
    & No resize && 3.16 && 3.53 & 5.67 && 12.87 & 2.35 & 20.22 \\
    & Frozen teacher && 3.10 && 3.36 & 5.24 && {\bf 8.11} & {\bf 2.38} & {\bf 13.90} \\
    & Default && {\bf 2.99} && {\bf 3.34} & {\bf 5.18} && 11.36 & 2.30 & 18.61 \\
    \hline
    \parbox[t]{2mm}{\multirow{4}{*}{\rotatebox[origin=c]{90}{Full}}} 
    & None && 2.67 && 3.18 & 4.60 && 4.10 & 2.02 & 12.95 \\
    & No resize && {\bf 2.51} && 3.14 & 4.48 && 3.53 & 2.02 & 11.13 \\
    & Frozen teacher && 2.66 && 3.04 & 4.24 && 2.99 & 1.99 & 9.70 \\
    & Default && 2.61 && {\bf 2.99} & {\bf 4.23} && {\bf 2.86} & {\bf 1.95} & {\bf 9.57} \\
    \bottomrule
    \end{tabular}}
    \caption{Losses on Sintel for zero flow, ground truth flow, and predicted flow}
    \label{table:loss-chart}
    \adjustbox{width=\linewidth}{
    \begin{tabular}{lcccccccccccccccc}
    \toprule
    &&\phantom{a} & \multicolumn{2}{c}{L1} & \phantom{a} & \multicolumn{2}{c}{SSIM} & \phantom{a} & \multicolumn{2}{c}{Census} & \phantom{a} & \multicolumn{1}{c}{SM} & \phantom{a}  & \multicolumn{2}{c}{Census + SM} \\
    \cmidrule{4-5} \cmidrule{7-8}  \cmidrule{10-11}  \cmidrule{13-13}   \cmidrule{15-16}  
    & Flow && noc & all && noc & all && noc & all && all && noc & all \\
    \midrule
    \parbox[t]{2mm}{\multirow{3}{*}{\rotatebox[origin=c]{90}{Clean}}}
    & Zero && .146 & .161 && .927 & .946 && 3.160 & 3.193 && 0. && 3.160 & 3.193 \\
    & GT && \textbf{.031} & .052 && \textbf{.191} & \textbf{.241} && \textbf{2.041} & \textbf{2.122} && .032 && \textbf{2.073} & \textbf{2.154} \\
    & UFlow && \textbf{.031} & \textbf{.042} && .203 & .247 && 2.06 & 2.130 && \textbf{.024} && 2.085 & \textbf{2.154} \\
    \midrule
    \parbox[t]{2mm}{\multirow{3}{*}{\rotatebox[origin=c]{90}{Final}}}
    & Zero && .126 & .142 && .731 & .751 && 3.037 & 3.075 && 0. && 3.037 & 3.075 \\
    & GT && .034 & .055 && .185 & .233 && 2.086 & 2.154 && .063 && 2.149 & 2.217 \\
    & UFlow && \textbf{.032} & \textbf{.037} && \textbf{.167} & \textbf{.226} && \textbf{2.044} & \textbf{2.091} && \textbf{.045} && \textbf{2.089} & \textbf{2.136} \\
    \bottomrule
    \end{tabular}}
    \vspace{-0.2cm}
\end{wraptable}

\noindent{\emph{Smoothness}:} Prior work suggests that photometric and smoothness losses taken together work better at higher resolutions~\cite{godard2019digging}. But our analysis of the smoothness loss alone shows an advantage of applying this loss at the resolution of flow estimation, rather than at the image resolution, in particular for Flying Chairs and Sintel (Table~\ref{table:smoothness_level}). Our results also show that first order smoothness works better on Chairs and Sintel while second order smoothness works better on KITTI (Table~\ref{table:smoothness_weights}). We see that context is important because in the minimal setting, the best second order smoothness weight for KITTI is 8, but in the full setting, it is 2. Comparing different edge-weights $\lambda$ (Eq.~\ref{eq:smooth}) in Table~\ref{table:smoothness_edge}, we see that nonzero edge-weights improve performance, particularly in the full setting. To our surprise, the simple strategy of only optimizing the Census loss and second order smoothness without edge-awareness, occlusion, or self-supervision (first row) produces performance on KITTI that improves on previous the state of the art.

\noindent{\emph{Self-Supervision}:} In Table~\ref{table:selfsup} we ablate the use of self-supervision and our proposed changes, and confirm that self-supervision on image crops is instrumental in achieving good results on KITTI, where errors are dominated by fast motion near the image edges. We also see that self-supervision is most effective when the image crop is resized as proposed by our method. Freezing the teacher network, as done in other works, seems to be important only when not using the other regularizing components. With these components in place, sharing the same model for both student and teacher appears to be beneficial.

\noindent{\emph{Loss Comparison to Ground Truth}:} Photometric loss functions used in unsupervised optical flow rely on the brightness consistency assumption: that pixel intensities in the camera image are invariant to motion in the world. But photometric consistency is an imperfect indicator of flow quality (e.g. in regions of shadows and specularity). To analyze this issue, we compute photometric and smoothness losses not only for the flow field produced by our model, but also for a flow field filled with zeros and for the ground truth flow. Table~\ref{table:loss-chart} shows that our model is able to achieve comparable or better photometric consistency (and overall loss) than the ground truth flow. This trend is more pronounced on Sintel Final, which we believe violates the consistency assumption more than Sintel Clean. This result suggests that the loss functions currently used may be a limiting factor in unsupervised methods.

\begin{wraptable}{r}{0.29\columnwidth}
    \vspace{-10.5pt}
    \caption{Resolution}
    \label{table:resolution}
    \adjustbox{width=\linewidth}{
    \begin{tabular}{llcccc@{}}
    \toprule
    & &\phantom{a}& \multicolumn{3}{c}{KITTI-15 \textit{train}} \\
    \cmidrule{4-6}
    & Resolution && all & noc & ER\%\\
    \midrule
    \parbox[t]{2mm}{\multirow{2}{*}{\rotatebox[origin=c]{90}{Min.}}}
    & 384$\times$1280 && 13.25 & 2.79 & 21.38 \\
    & 640$\times$640 && 12.91 & 2.42 & 21.17 \\
    \hline
    \parbox[t]{2mm}{\multirow{2}{*}{\rotatebox[origin=c]{90}{Full}}} 
    & 384$\times$1280 && 3.80 & 2.13 & 10.88 \\
    & 640$\times$640 && \bf{2.93} & \bf{1.96} & \bf{9.61} \\
    \bottomrule
    \end{tabular}}
\end{wraptable}

\noindent{\emph{Resolution}:} Table~\ref{table:resolution} shows, perhaps surprisingly, that estimating flow at a different resolution and aspect ratio can substantially improve performance on KITTI-15 (2.93 vs. 3.80), presumably because the motion field in this dataset is dominated by horizontal motion. We have not observed this effect in other datasets.

\begin{wraptable}{r}{0.45\columnwidth}
    \vspace{-6.5pt}
    \caption{Data augmentation. F: image flipping up/down and left/right (not used for KITTI), C: color augmentation} 
    \label{table:data_augmentation}
    \adjustbox{width=\linewidth}{
    \begin{tabular}{lcccccccccccc}
    \toprule
    & \phantom{a} && &\phantom{a}& Chairs & \phantom{a} & \multicolumn{2}{c}{Sintel \textit{train}} & \phantom{a} & \multicolumn{3}{c}{KITTI-15 \textit{train}} \\
    \cmidrule{6-6} \cmidrule{8-9} \cmidrule{11-13} 
    && F & C && \textit{test} && Clean & Final && all & noc & ER\%\\
    \midrule
    \parbox[t]{2mm}{\multirow{3}{*}{\rotatebox[origin=c]{90}{Minimal}}}
    && -- & -- && {\bf 3.47} && 4.39 & {\bf 6.56} && 13.27 & 2.56 & 22.13 \\
    && -- & \checkmark && 3.56 && 4.38 & 6.58 && {\bf 13.07} & {\bf 2.47} & {\bf 21.21} \\
    && \checkmark & -- && 3.49 && 4.23 & 6.73 && -- & -- & -- \\
    && \checkmark & \checkmark && 3.58 && {\bf 4.20} & 6.80 && -- & -- & -- \\
    \hline
    \parbox[t]{2mm}{\multirow{3}{*}{\rotatebox[origin=c]{90}{Full}}} 
    && -- & -- && 2.53 && 3.84 & 5.14 && 3.06 & 2.03 & 9.82 \\
    && -- & \checkmark && 2.61 && 3.78 & 5.23 && {\bf 2.94} & {\bf 1.98} & {\bf 9.65} \\
    && \checkmark & -- && 2.57 && 3.02 & 4.22 && -- & -- & -- \\
    && \checkmark & \checkmark && {\bf 2.55} && {\bf 3.00} & {\bf 4.18} && -- & -- & -- \\
    \bottomrule
    \end{tabular}}
    \caption{Pretraining on Chairs}
    \label{table:pretraining}
    \adjustbox{width=\linewidth}{
    \begin{tabular}{lcccccccc@{}}
    \toprule
    & Pretraining & \phantom{a} & \multicolumn{2}{c}{Sintel \textit{train}} & \phantom{a} & \multicolumn{3}{c}{KITTI-15 \textit{train}} \\
    \cmidrule{4-5} \cmidrule{7-9} 
    & on Chairs && Clean & Final && all & noc & ER\%\\
    \midrule
    \parbox[t]{2mm}{\multirow{2}{*}{\rotatebox[origin=c]{90}{Min.}}}
    & -- && 4.41 & 7.53 && {\bf 12.93} & {\bf 2.44} & 21.24 \\
    & \checkmark && {\bf 4.20} & {\bf 6.80} && 13.07 & 2.47 & {\bf 21.21} \\
    \hline
    \parbox[t]{2mm}{\multirow{2}{*}{\rotatebox[origin=c]{90}{Full}}} 
    & -- && 3.38 & 4.81 && 3.08 & 2.04 & 10.00 \\
    & \checkmark && {\bf 3.00} & {\bf 4.18} && {\bf 2.94} & {\bf 1.98} & {\bf 9.65} \\
    \bottomrule
    \end{tabular}}
    \caption{Gradient steps (S) and batch size (B)}
    \label{table:steps_and_bs}
    \adjustbox{width=\linewidth}{
    \begin{tabular}{llccccccccccc}
    \toprule
    &\phantom{a}&&&\phantom{a}& Chairs & \phantom{a} & \multicolumn{2}{c}{Sintel \textit{train}} & \phantom{a} & \multicolumn{3}{c}{KITTI-15 \textit{train}} \\
    \cmidrule{6-6} \cmidrule{8-9} \cmidrule{11-13} 
    && S & B && \textit{test} && Clean & Final && all & noc & ER\%\\
    \midrule
    \parbox[t]{2mm}{\multirow{2}{*}{\rotatebox[origin=c]{90}{\textit{test}}}}
    &&60K & 32 && \{3.16\} && 3.04 & 4.23 && 2.92 & {\bf 1.96} & 9.71 \\
    &&1.2M & 1 && \{2.82\} && {\bf 3.01} & {\bf 4.09} && {\bf 2.84} & {\bf 1.96} & {\bf 9.39} \\
    \parbox[t]{2mm}{\multirow{2}{*}{\rotatebox[origin=c]{90}{\textit{train}}}}
    &&60K & 32 && 2.57 && \{2.47\} & \{3.92\} && \{2.74\} & \{1.87\} & \{9.04\} \\
    &&1.2M & 1 && {\bf 2.55} && \{2.50\} & \{3.39\} && \{2.71\} & \{1.88\} & \{9.05\} \\
    \bottomrule
    \end{tabular}}
    \vspace{-1cm}
\end{wraptable}

\noindent{\emph{Data Augmentation}:} Table~\ref{table:data_augmentation} evaluates the importance of color augmentation (color channel swapping and hue randomization) for all domains, as  well as image flipping for Sintel. The results show that both augmentation techniques improve performance, in particular image flipping for Sintel (which is a much smaller dataset than Chairs or KITTI).

\noindent{\emph{Pretraining}:} Pretraining is a common strategy in supervised~\cite{FlowNet,Sun2018PWCNet} and unsupervised~\cite{DDFlow,Zhong2019UnsupervisedDE} optical flow. The results in Table~\ref{table:pretraining} confirm that pretraining on Chairs improves performance on Sintel and KITTI.

\noindent{\emph{Gradient Steps and Batch Size}:} All experiments up to this point have trained the model for 60K steps at a batch size of 32. Table~\ref{table:steps_and_bs} shows a comparison to another training regime that trains longer with smaller batches, which consistently improves performance. We use this regime for our comparison to other published methods.

\begin{table}[t]
    \centering
    \caption{Our model (yellow) compared to state of the art. Supervised models in gray fine-tune on their evaluation domain, which is often not possible in practice. Braces indicate models whose training set includes its evaluation set, and so are not comparable: ``()'' trained on the labeled evaluation set, ``\{\}'' trained on the unlabeled evaluation set, and ``[]'' trained on data related to the evaluation set (e.g. $<5$ frames away in KITTI, or having the same content in Sintel). The best unsupervised and supervised (without finetuning) results are in bold. Methods that use additional modalities are denoted with MDM: mono depth/motion, SDM: stereo depth/motion, MF: multi-frame flow}
    \vspace{-3.5mm}
    \label{tab:main}
    \adjustbox{max width=0.97\textwidth}{
	\begin{tabular}{lcl cc c cc c cc c cccc}
	    \toprule
		 && & \multicolumn{2}{c}{Sintel Clean~\cite{ButlerECCV2012}} & \phantom{a} & \multicolumn{2}{c}{Sintel Final~\cite{ButlerECCV2012}} & \phantom{a} & \multicolumn{2}{c}{KITTI 2012~\cite{Geiger2012CVPR}} & \phantom{a} & \multicolumn{4}{c}{KITTI 2015~\cite{KITTI2015}} \\
		 \cmidrule{4-5} \cmidrule{7-8} \cmidrule{10-11} \cmidrule{13-16}     
        && & \multicolumn{2}{c}{EPE} && \multicolumn{2}{c}{EPE} && \multicolumn{2}{c}{EPE} && EPE & EPE (noc) & \multicolumn{2}{c}{ER in \%}\\
        &&Method& \textit{train} & \textit{test} && \textit{train} & \textit{test} && \textit{train} & \textit{test} && \textit{train} & \textit{train} & \textit{train} & \textit{test}\\
        \midrule
        \parbox[t]{2mm}{\multirow{7}{*}{\rotatebox[origin=c]{90}{\small{Supervised}}}}
        & \cellcolor{gray}(A) & \cellcolor{gray}FlowNet2-ft \cite{Flownet2} & \cellcolor{gray}(1.45) & \cellcolor{gray}4.16 &\cellcolor{gray}& \cellcolor{gray}(2.01) & \cellcolor{gray}5.74 &\cellcolor{gray}& \cellcolor{gray}(1.28) & \cellcolor{gray}1.8 &\cellcolor{gray}& \cellcolor{gray}(2.30) & \cellcolor{gray}-- & \cellcolor{gray}(8.61) & \cellcolor{gray}11.48 \\
        & \cellcolor{gray}(B) & \cellcolor{gray}PWC-Net-ft \cite{Sun2018PWCNet} & \cellcolor{gray}(1.70) & \cellcolor{gray}3.86 &\cellcolor{gray}& \cellcolor{gray}(2.21) & \cellcolor{gray}5.13 &\cellcolor{gray}& \cellcolor{gray}(1.45) & \cellcolor{gray}1.7 &\cellcolor{gray}& \cellcolor{gray}(2.16) & \cellcolor{gray}-- & \cellcolor{gray}(9.80) &\cellcolor{gray}9.60 \\
        & \cellcolor{gray}(C) & \cellcolor{gray}SelFlow-ft \cite{SelFlow} & \cellcolor{gray}(1.68) & \cellcolor{gray}[3.74] &\cellcolor{gray}& \cellcolor{gray}(1.77) & \cellcolor{gray}\{4.26\} &\cellcolor{gray}& \cellcolor{gray}(0.76) & \cellcolor{gray}1.5 &\cellcolor{gray}& \cellcolor{gray}(1.18) & \cellcolor{gray}-- & \cellcolor{gray}-- & \cellcolor{gray}8.42 \\ 
        & \cellcolor{gray}(D) & \cellcolor{gray}VCN-ft \cite{yang2019volumetric}  & \cellcolor{gray}(1.66) & \cellcolor{gray}2.81 &\cellcolor{gray}& \cellcolor{gray}(2.24) & \cellcolor{gray}4.40 &\cellcolor{gray}& \cellcolor{gray}-- & \cellcolor{gray}-- &\cellcolor{gray}& \cellcolor{gray}(1.16) & \cellcolor{gray}-- &  \cellcolor{gray}(4.10) & \cellcolor{gray}6.30 \\
        & (E)& FlowNet2 \cite{Flownet2} & {\bf 2.02} & {\bf 3.96} && {\bf 3.14} & {\bf 6.02} && {\bf 4.09} & -- && 9.84 & -- & 28.20 & -- \\
        & (F)& PWC-Net \cite{Sun2018PWCNet} & 2.55 & -- && 3.93 & -- && 4.14 & -- && 10.35 & -- & 33.67 & -- \\
        & (G)& VCN \cite{yang2019volumetric} & 2.21 & -- && 3.62 & -- && -- & -- && {\bf 8.36} & -- & {\bf 25.10} & -- \\
        \midrule
        \parbox[t]{2mm}{\multirow{14}{*}{\rotatebox[origin=c]{90}{Unsupervised}}}
        &(H)& Back2Basics \cite{jjyu2016unsupflow} & -- & -- && -- & -- && 11.30 & 9.9 && -- & -- & -- & -- \\ %
        &(I)& DSTFlow \cite{ren2017unsupervised} & \{6.16\} & 10.41 && \{7.38\} & 11.28 && [10.43] & 12.4 && [16.79] & [6.96] & [36.00] & [39.00]\\ %
        &(J)& OAFlow \cite{wang2018occlusion}& \{4.03\} & 7.95 && \{5.95\} & 9.15 && [3.55] & [4.2]  && [8.88] & -- & -- & [31.20] \\ %
        &(K)& UnFlow \cite{meister2018unflow} & -- & -- && 7.91 & 10.21 && 3.29 & -- && 8.10 &--& 23.27 & --\\ %
        &(L)& GeoNet \cite{yin2018geonet}~$^{\text{(MDM)}}$ &--&--&&--&-- && --&--&& 10.81 & 8.05 &-- & --\\  %
        &(M)& DF-Net \cite{zou2018dfnet}~$^{\text{(MDM)}}$& -- & -- && -- & -- && 3.54 & 4.4 && \{8.98\} & -- & \{26.01\} & \{25.70\}\\ %
        &(N)& CCFlow \cite{ranjan2019cvpr}~$^{\text{(MDM)}}$ &--&--&&--&-- &&--&-- && 5.66 &-- & 20.93 & 25.27 \\ %
        &(O)& MFOccFlow \cite{Janai2018ECCV}~$^{\text{(MF)}}$ & \{3.89\} & 7.23 && \{5.52\} & 8.81 && --& --&& [6.59] & [3.22] & -- & 22.94\\ %
        &(P) & UnOS \cite{wang2018unos}~$^{\text{(SDM)}}$ &--&--&&--&-- && 1.64 & {\bf 1.8} && 5.58 & -- & -- & 18.00\\ %
        &(Q)& EPIFlow \cite{Zhong2019UnsupervisedDE} & 3.94 & 7.00 && 5.08 & 8.51 && 2.61 & 3.4 && 5.56 & 2.56 & -- & 16.95 \\ %
        &(R)& DDFlow \cite{DDFlow} & \{2.92\} & 6.18 && \{3.98\} & 7.40 && [2.35] & 3.0 && [5.72] & [2.73] & -- & 14.29 \\ %
        &(S)& SelFlow \cite{SelFlow}~$^{\text{(MF)}}$ & [2.88] & [6.56] && \{3.87\} & \{6.57\} && [1.69] & 2.2 && [4.84] & [2.40] & -- & 14.19  \\ %
        \rowcolor{lightyellow} & (T) & UFlow-test & {\bf 3.01} & -- && {\bf 4.09} & -- && {\bf 1.58} & -- && {\bf 2.84} & {\bf 1.96} & {\bf 9.39} & -- \\
        \rowcolor{lightyellow} & (U) & UFlow-train & \{2.50\} & {\bf 5.21} && \{3.39\} & {\bf 6.50} && 1.68 & 1.9 && \{2.71\} & \{1.88\} & \{9.05\} & {\bf 11.13} \\
        \bottomrule
	\end{tabular}}
\end{table}

\begin{figure}[t]
\centering
\resizebox{\linewidth}{!}{
\begin{tabular}{@{}c@{\,\,}c@{\resultsspace}c@{\resultsspace}c@{\resultsspace}c@{\resultsspace}c@{\resultsspace}c@{}}
\rotatebox{90}{\footnotesize \,2015} & \includegraphics[width = \resultswidth]{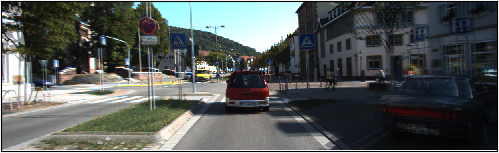} & 
\includegraphics[width = \resultswidth]{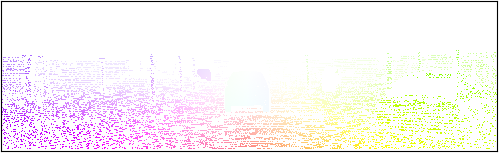} &
\includegraphics[width = \resultswidth]{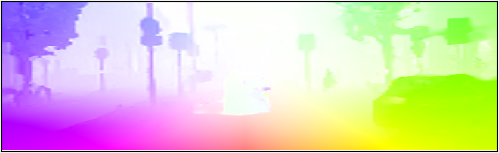} &
\includegraphics[width = \resultswidth]{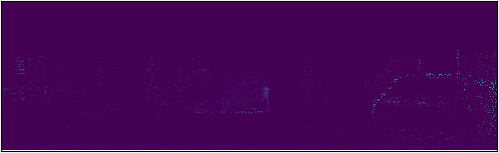} &
\includegraphics[width = \resultswidth]{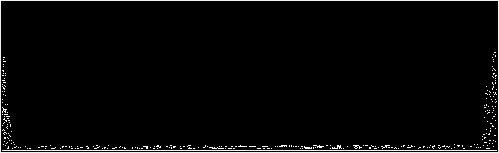} & 
\includegraphics[width = \resultswidth]{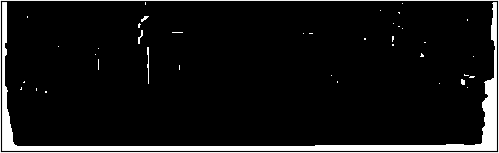} \\
\rotatebox{90}{\footnotesize \,\,KITTI} & \includegraphics[width = \resultswidth]{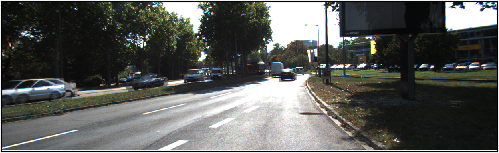} &
\includegraphics[width = \resultswidth]{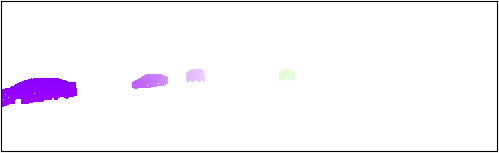} & 
\includegraphics[width = \resultswidth]{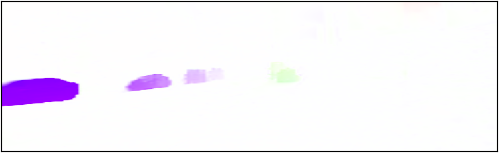} &
\includegraphics[width = \resultswidth]{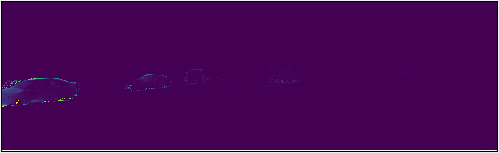} &
\includegraphics[width = \resultswidth]{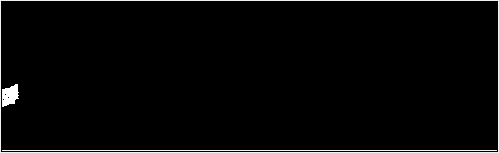} &
\includegraphics[width = \resultswidth]{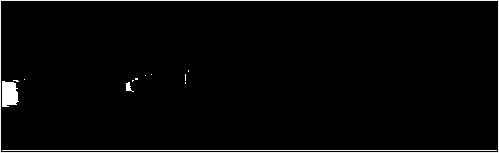} \\
\cmidrule(l{0pt}r{0pt}){1-7}
\rotatebox{90}{\footnotesize Clean} & \includegraphics[width = \resultswidth]{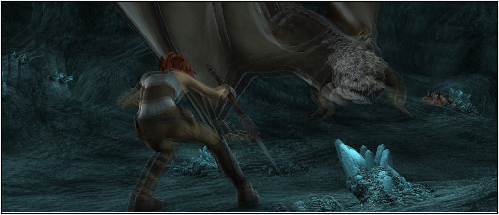} &
\includegraphics[width = \resultswidth]{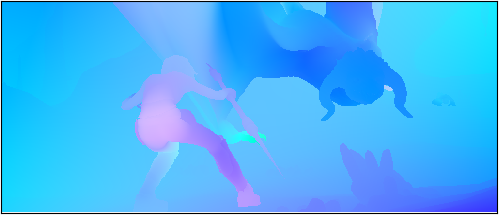} &
\includegraphics[width = \resultswidth]{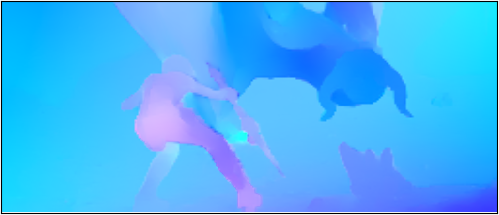} &
\includegraphics[width = \resultswidth]{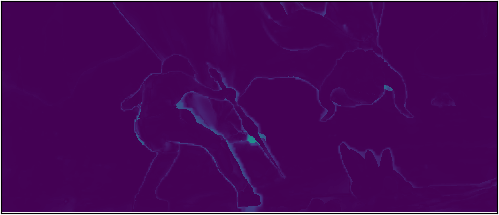} &
\includegraphics[width = \resultswidth]{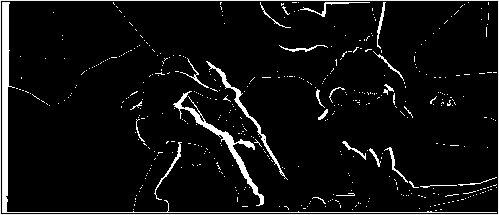} &
\includegraphics[width = \resultswidth]{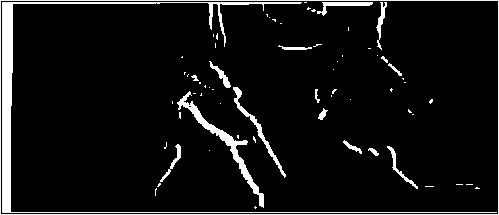} \\
\rotatebox{90}{\footnotesize \,\,\,\,\,Sintel} & \includegraphics[width = \resultswidth]{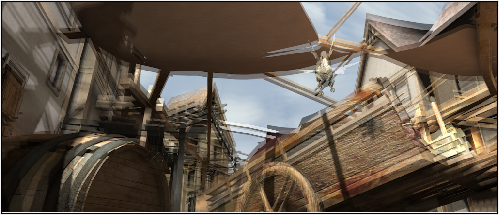} &
\includegraphics[width = \resultswidth]{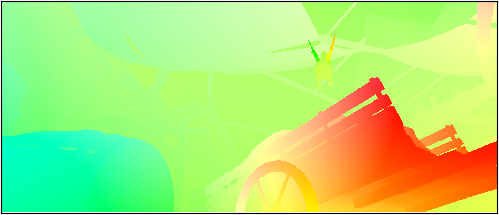} &
\includegraphics[width = \resultswidth]{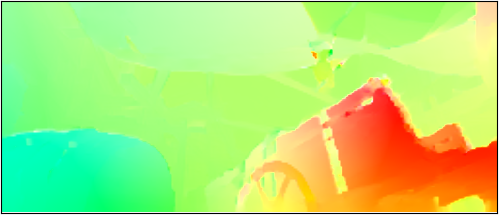} &
\includegraphics[width = \resultswidth]{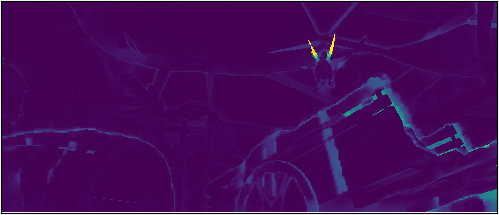} &
\includegraphics[width = \resultswidth]{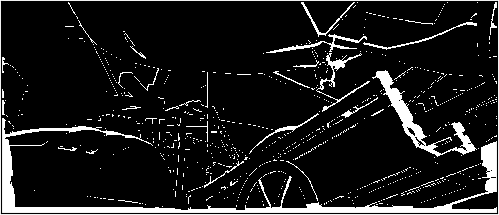} &
\includegraphics[width = \resultswidth]{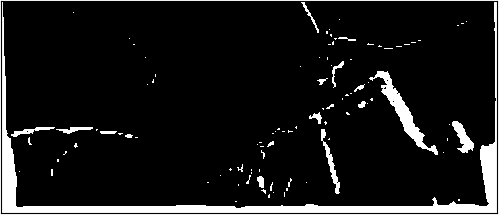} \\
\cmidrule(l{0pt}r{0pt}){1-7}
\rotatebox{90}{\footnotesize Final} & \includegraphics[width = \resultswidth]{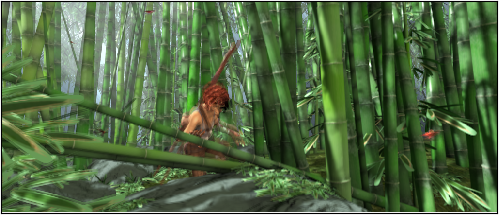} &
\includegraphics[width = \resultswidth]{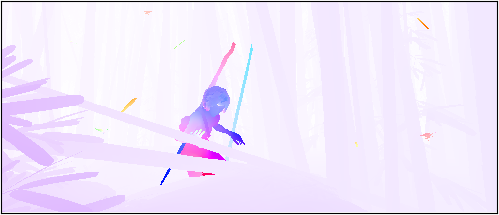} &
\includegraphics[width = \resultswidth]{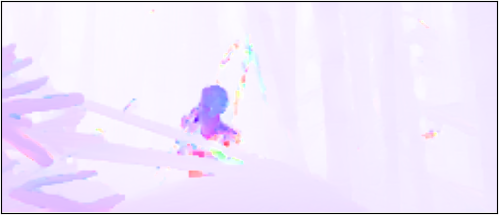} &
\includegraphics[width = \resultswidth]{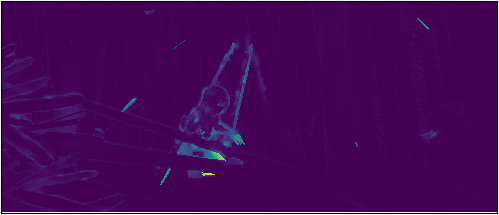} &
\includegraphics[width = \resultswidth]{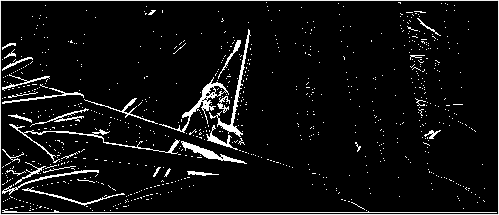} &
\includegraphics[width = \resultswidth]{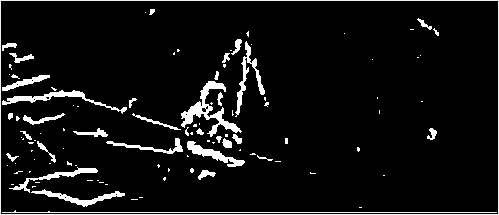} \\
\rotatebox{90}{\footnotesize \,\,\,\,\,Sintel} & \includegraphics[width = \resultswidth]{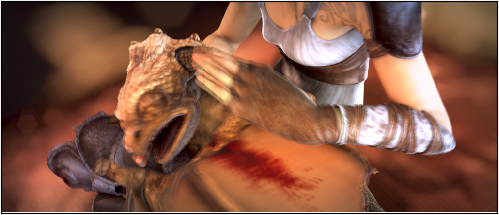} &
\includegraphics[width = \resultswidth]{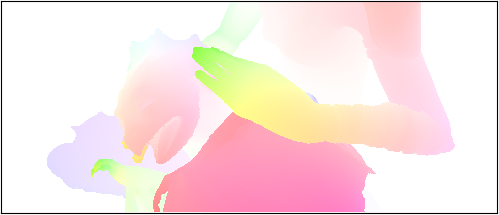} &
\includegraphics[width = \resultswidth]{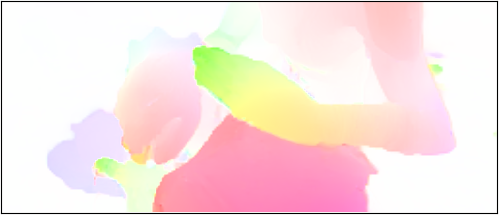} &
\includegraphics[width = \resultswidth]{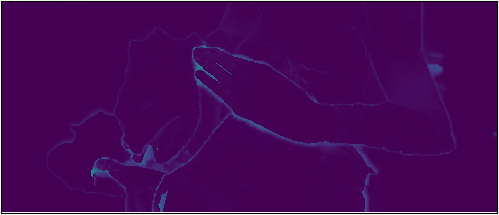} &
\includegraphics[width = \resultswidth]{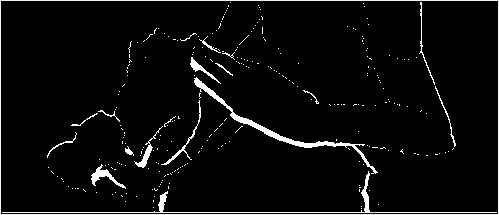} &
\includegraphics[width = \resultswidth]{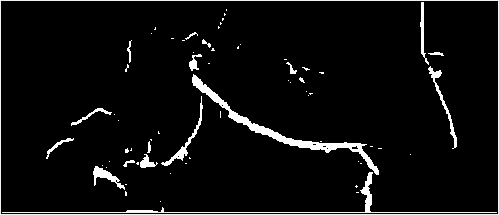} \\
& {\footnotesize Input RGB} & {\footnotesize True Flow} & {\footnotesize Predicted Flow} & {\footnotesize Endpoint Error} & {\footnotesize True Occlusions} & {\footnotesize Predicted Occlusions}
\end{tabular}
}
\caption{
Results for our model on random examples not seen during training taken from KITTI 2015 and Sintel Final.
These qualitative results show the model's ability to estimate fast motions, relatively fine details, and substantial occlusions
}
\label{fig:results}
\end{figure}

\subsubsection{Comparison to State of the Art:}

We show qualitative results in Figure~\ref{fig:results} and quantitatively evaluate our model trained on KITTI and Sintel data in the corresponding benchmarks in Table~\ref{tab:main}, where we compare against state-of-the-art techniques for unsupervised and supervised optical flow. Results not reported by prior work are indicated with ``--''.

Among unsupervised approaches (H-U), our model sets a new state of the art for Sintel Clean (5.21 vs. 6.18), Sintel Final (6.50 vs. 7.40), and KITTI-15 (11.13\% vs. 14.19\%) -- where, for a lack of comparability, we had to disregard results in braces that came from (partially) training on the test set. UFlow is only outperformed (1.8 vs. 1.9) on KITTI-12, which does not include moving objects, by a stereo-depth and motion based approach (P).

The top-performing supervised models finetuned on data from the evaluation domain (models A-D) do outperform our unsupervised model, as one may expect. But on KITTI-15, our model \emph{performs on par with the supervised FlowNet2}. Of course, fine-tuning on the domain is only possible because the KITTI training data also contains ground-truth flow, which we ignore but which supervised techniques require. This sort of supervision is hard to obtain (KITTI being virtually the only non-synthetic dataset with this information), which demonstrates the value of unsupervised flow techniques such as ours. Without access to the ground truth labels of the test domain, our unsupervised method compares more favorably to its supervised counterparts, significantly outperforming them on KITTI. Our final experiment analyses cross-domain generalization in more detail.

\pagebreak

\begin{wraptable}{r}{0.5\columnwidth}
    \vspace{-10.5pt}
    \caption{Generalization across datasets. Performance when training on one dataset and testing on different one (gray if same)}
    \label{table:crossdata}
    \adjustbox{width=\linewidth}{
    \begin{tabular}{lllccccccccc}
    \toprule
    &&&\phantom{a}& Chairs & \phantom{a} & \multicolumn{2}{c}{Sintel \textit{train}} & \phantom{a} & \multicolumn{3}{c}{KITTI-15 \textit{train}} \\
    \cmidrule{5-5} \cmidrule{7-8} \cmidrule{10-12} 
    && Method && \textit{test} && Clean & Final && all & noc & ER\%\\
    \midrule
    \parbox[t]{2mm}{\multirow{4}{*}{\rotatebox[origin=c]{90}{Train on}}} &
    \parbox[t]{2mm}{\multirow{4}{*}{\rotatebox[origin=c]{90}{Chairs}}}
    & PWC-Net~\cite{Sun2018PWCNet} && \cellcolor{gray} 2.00 && {\bf 3.33} & 4.59 && 13.20 & -- & 41.79 \\
    && DDFlow~\cite{DDFlow} &&\cellcolor{gray} 2.97 && 4.83 & 4.85 && 17.26 & -- & -- \\
    &&  UFlow-test && \cellcolor{gray} \{2.82\} && 4.36 & 5.12 && 15.68 & 7.96 & 32.69 \\
    &&  UFlow-train && \cellcolor{gray} 2.55 && 3.43 & {\bf 4.17} && {\bf 11.27} & {\bf 5.66} & {\bf 30.31} \\
    \hline
    \parbox[t]{2mm}{\multirow{4}{*}{\rotatebox[origin=c]{90}{Train on}}}&
    \parbox[t]{2mm}{\multirow{4}{*}{\rotatebox[origin=c]{90}{Sintel}}} 
    & PWC-Net~\cite{Sun2018PWCNet} && 3.69 && \cellcolor{gray} (1.86) & \cellcolor{gray} (2.31) && 10.52 & -- & 30.49 \\
    && DDFlow~\cite{DDFlow} && 3.46 && \cellcolor{gray} \{2.92\} & \cellcolor{gray} \{3.98\} && 12.69 & -- & -- \\
    &&  UFlow-test && 3.39 && \cellcolor{gray} 3.01 & \cellcolor{gray} 4.09 && {\bf 7.67} & {\bf 3.77} & {\bf 17.41} \\
    &&  UFlow-train && {\bf 3.25} && \cellcolor{gray} \{2.50\} & \cellcolor{gray} \{3.39\} && 9.40 & 4.53 & 20.02 \\
    \hline
    \parbox[t]{2mm}{\multirow{3}{*}{\rotatebox[origin=c]{90}{Train on}}} &
    \parbox[t]{2mm}{\multirow{3}{*}{\rotatebox[origin=c]{90}{KITTI}}} 
    & DDFlow~\cite{DDFlow} && 6.35 && 6.20 & 7.08 && \cellcolor{gray} [5.72] &\cellcolor{gray} -- & \cellcolor{gray} -- \\
    &&  UFlow-test && 5.25 && 6.34 & 7.01 && \cellcolor{gray} 2.84 & \cellcolor{gray} 1.96 & \cellcolor{gray} 9.39 \\
    &&  UFlow-train && {\bf 5.05} && {\bf 5.58} & {\bf 6.31} && \cellcolor{gray} \{2.71\} & \cellcolor{gray} \{1.88\} & \cellcolor{gray} \{9.05\} \\
    \bottomrule
    \end{tabular}}
\end{wraptable}

Table~\ref{table:crossdata} evaluates out-of-domain generalization by training and evaluating models across three datasets. While performance is best when training and test data are from the same domain, our model shows good generalization. It consistently outperforms DDFlow and it outperforms the supervised PWC-Net in all but one generalization task (training on Chairs and testing on Sintel Clean).

\section{Conclusion}

We have presented a study into what matters in unsupervised optical flow that systematically analyzes, compares, and improves a set of key components. This study results in a range of novel observations about these components and their interactions, from which we integrate the best components and improvements into a unified framework for unsupervised optical flow. Our resulting UFlow model
substantially outperforms the state of the art among unsupervised methods and performs on par with the supervised FlowNet2 on the challenging KITTI 2015 benchmark, despite not using any labels. In addition to its strong performance, our method is also significantly simpler than many related approaches, which we hope will make it useful as a starting point for further research into unsupervised optical flow. Our code is available at \url{https://github.com/google-research/google-research/tree/master/uflow}.
	
\clearpage

{\small
\bibliographystyle{ieee_fullname}
\bibliography{references}
}

\end{document}

%% file: macros.tex
\def\rot#1{\rotatebox{90}{#1}}

\newcommand{\abs}[1]{\left\lvert#1\right\rvert}
\newcommand{\mean}[1]{\frac{1}{n} \sum #1}

\newcommand{\lossfun}[1]{\mathcal{L}_\mathit{#1}}

\newcommand{\level}{\ell}

\newcommand{\etal}{\textit{et al}. }